\pdfoutput=1
\documentclass[11pt]{article}

\usepackage[utf8]{inputenc}
\usepackage{mathptmx}
\usepackage[margin=1in]{geometry}
\usepackage{booktabs,graphicx,amsmath,amssymb,xcolor,array}
\usepackage[numbers,sort&compress]{natbib}
\usepackage{microtype}
\usepackage{authblk}
\usepackage{enumitem}
\usepackage{placeins}
\usepackage{url}
\definecolor{linkblue}{rgb}{0.21,0.49,0.74}
\usepackage[breaklinks,colorlinks,allcolors=linkblue]{hyperref}
\graphicspath{{figures/}}
\title{Leakage-Free Is Not Confounder-Free:\\
What a Patient-Level Split Cannot Certify About a Medical Imaging Benchmark}

\author[1]{Ignazio Gallo\thanks{Corresponding author: \texttt{ignazio.gallo@uninsubria.it}}}
\author[1]{Mattia Gatti}
\author[1]{Leonardo Sgroi}
\author[2]{Darwin Castillo}
\author[2]{Yuliana Jim\'enez-Gaona}
\author[3]{Nicola Landro}
\affil[1]{Department of Theoretical and Applied Sciences, University of Insubria, Varese, Italy}
\affil[2]{Departamento de Química, Facultad de Ciencias Exactas y Naturales, Universidad Técnica Particular de Loja, Loja, Ecuador}
\affil[3]{Joint Research Centre, Ispra, Italy}

\date{}

\begin{document}
\maketitle

\begin{abstract}
Patient-level splitting is the standard safeguard against subject leakage in medical
imaging evaluation, but it cannot remove a confounder that every fold inherits: an
acquisition variable aligned with the diagnosis across the whole cohort. We present a
five-question audit of a candidate confounder: how well it predicts the diagnosis,
whether it can be decoded from the images, whether diagnostic scores depend on it,
whether the image adds predictive value beyond it and how much discrimination survives
once it is controlled. We apply it to DMR-IR, the public dataset on which most
machine-learning results in breast thermography rest. On 236 patients under a strict
patient-disjoint protocol, the acquisition date alone predicts malignancy at AUROC
0.980, above every imaging arm we evaluated (best 0.949), because benign and malignant
patients were recruited in different years. The acquisition year is decodable from the
pixels among patients who share a diagnosis, largely through a global thermal offset
outside the breast, yet only one arm, the one pooling every frame of a visit,
demonstrably carries it into its diagnostic score. No arm improves on the date on paired
folds, and once the date is controlled, discrimination either collapses or survives only
on comparisons the benchmark cannot attribute. The dataset contains only eight
cross-label same-date pairs, from ten patients, which limits reliable evaluation
within recorded acquisition-date groups. A second, independently acquired cohort, whose
dominant channel is patient age, reproduces every finding except legibility, from a
confounder that is clinically legitimate. Mitigation may suppress such a confounder, but without cross-label
acquisition support no evaluation on the benchmark can certify that what remains is
pathology.
\end{abstract}


\section{Introduction}
A patient-level split has become the marker of a trustworthy medical imaging
evaluation: if no patient contributes images to both sides of the partition, the
estimate is presumed to measure the model rather than the leak. This paper is
about what that presumption misses. A split removes \emph{duplicated patients};
it does nothing to a confounder that every fold inherits, because a property of
how the data were acquired (when, where, on which device) can be aligned
with the diagnosis in the dataset as a whole, and no partition of patients
disturbs that alignment. That acquisition correlates such as site, scanner,
markers and burned-in annotations can act as shortcuts is well
documented~\cite{zech2018variable,degrave2021covid,hill2024shortcutting}; what is less
often appreciated is that the patient-level split, routinely reported as the evidence
that an evaluation is leakage-free, offers no protection against them.

We study this on DMR-IR~\cite{silva2014dmr}, the public breast-thermography
benchmark on which most machine-learning results in the field rest
(Fig.~\ref{fig:audit}). Under a patient-disjoint protocol, the acquisition date
alone achieves a higher AUROC than every imaging arm we evaluated. The acquisition
era is decodable from the images within diagnosis groups, and the diagnostic scores
of one configuration, ConvNeXt-T over all frames of a visit, show a detectable
within-diagnosis association with it. No incremental predictive value beyond the
date is demonstrated under the combinations tested. Restricting the evaluation
reduces the discrimination of the frontal model, while the residual discrimination
of the all-frames models remains compatible with more than one explanation.
Because 83 of the 86 acquisition-date groups contain a single class and the cohort
holds only eight cross-label same-date pairs, these analyses cannot reliably
separate disease-related from acquisition-related information.

The contribution is not that DMR-IR is flawed; it is the demonstration, on a
real and consequential benchmark, that \textbf{patient-level evaluation is
necessary but not sufficient}. Concretely, the paper contributes:
\begin{itemize}[leftmargin=*,itemsep=2pt,topsep=4pt]
\item a five-question audit (Q1--Q5: confounder strength, legibility, exploitation,
incremental value and controlled discrimination; \S\ref{sec:setup}) that needs only
existing images, metadata and model predictions, including a conditional AUROC
reported beside the same statistic computed for the candidate confounder
(\S\ref{sec:controlled});
\item its application to DMR-IR, answered one question per subsection in
\S\ref{sec:results};
\item the measured distinction between a confounder being \emph{legible} in the pixels
and being \emph{exploited} by a diagnostic model, which on this benchmark diverge sharply
(\S\ref{sec:legibility});
\item an assessment of cross-label acquisition support that identifies which
controlled comparisons the observed data can support and where additional data or
assumptions are required (\S\ref{sec:repair}).
\end{itemize}
A second thermography cohort, whose dominant non-image variable is age, a clinically
relevant predictor, provides an additional application of the audit; validation
across imaging modalities and against confounding mechanisms of known strength
remains future work.

\section{Related work}
\textbf{Leakage.} Leakage (label information reaching the model by a path the
deployment setting will not provide) has a general formulation and taxonomy in
data mining~\cite{kaufman2012leakage}, of which subject-level leakage from
image-level splitting of multi-image datasets is the form medical imaging
inherits; across 17 scientific fields it is a documented and widespread source
of irreproducible results~\cite{kapoor2023leakage}. Our split-protocol ladder
quantifies it here at $+0.046$ AUROC, with
contamination measured rather than asserted, and separates it from a comparably
sized effect ($+0.041$) of scoring patients rather than frames
(Appendix~\ref{sec:splitunit}).

\textbf{Shortcut learning and batch effects.} Medical imaging models routinely
key on acquisition correlates rather than pathology: hospital-system and scanner
markers in chest radiographs~\cite{zech2018variable}, dataset provenance in
COVID-19 classifiers~\cite{degrave2021covid}, the phenomenon catalogued in
general as shortcut learning~\cite{geirhos2020shortcut}; recent work shows how
little of the shortcut need be clinical (knee radiographs predict whether a
patient drinks beer, through site and scanner markers~\cite{hill2024shortcutting})
and that the resulting overestimation of reported performance is
systematic~\cite{ongly2024shortcut}. High-throughput biology reached the same
conclusion earlier and treats batch structure as a first-class design
problem~\cite{leek2010batch}; aggregate metrics conceal clinically meaningful
failure on unlabelled subgroups~\cite{oakdenrayner2020hidden}; dataset bias has
a causal formulation~\cite{jones2024causal} and imaging benchmarks a standing
audit of the artefacts they carry~\cite{jimenezsanchez2025picture}. What is rarely
available is a protocol under which confounder strength, pixel-level legibility,
model exploitation, incremental value and controlled re-evaluation are all
measured on the same folds; that is what this paper provides.

\textbf{Availability and predictivity.} Hermann et al.~\cite{hermann2024foundations}
distinguish two properties of a feature, its \emph{predictivity} (how reliably it indicates
the label) and its \emph{availability} (how easily a model extracts it), and show on
synthetic data, Waterbirds and CelebA that availability can override predictivity in
deciding which feature a network relies on. Our audit addresses related questions
on an observational medical-imaging benchmark. Q1 measures the predictive
association of a candidate non-image variable with the diagnosis. Q2 tests whether
that variable can be decoded from the images within diagnosis strata; this is an
operational measure of recoverable information rather than a direct estimate of
availability in their sense. Q3 measures the dependence of the diagnostic score on
that variable within diagnosis strata; such dependence is evidence of an
association in model outputs, but does not by itself identify causal reliance. Q4
and Q5 assess incremental predictive value and discrimination under restricted
comparisons. Our contribution is the joint application of these diagnostics under
one patient-disjoint protocol, together with an assessment of whether the benchmark
holds the cross-label acquisition support needed to attribute what survives
control.

\textbf{Multimodal thermography.} Recent work fuses the thermogram with the
clinical record; one study~\cite{pinto2025multimodal} reports 0.918 accuracy
from the record alone. Our attribution of that channel, including a released-file
defect by which two age columns reconstruct the acquisition calendar (their
difference correlates with true elapsed time at Pearson 0.990 and the pair
scores 0.946), is in Appendix~\ref{sec:metadata}.

\section{Setup}
\label{sec:setup}
\begin{figure}[t]
\centering
\includegraphics[width=\linewidth]{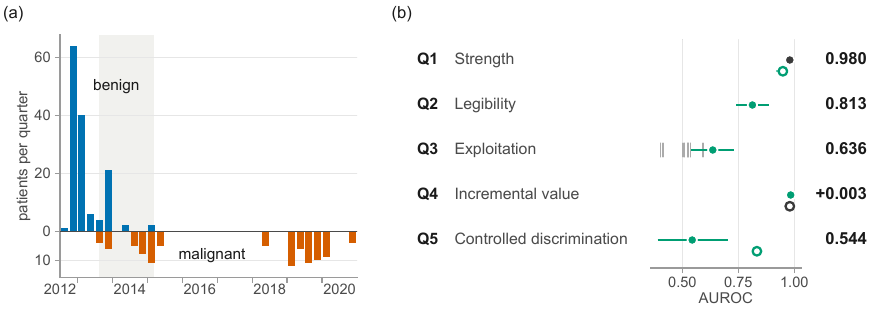}
\caption{\textbf{Leakage-free is not confounder-free.} (a)~The 236 DMR-IR
patients (140 benign, 96 malignant) by quarter of acquisition, benign above the
axis and malignant below. The shaded date-overlap window, from the first malignant to the
last benign acquisition, is the only stretch of time both classes occupy (60 patients),
and 83 of the 86 acquisition-date groups contain a single class. A patient-level
split partitions these patients; no partition of them changes when they were
photographed, so every fold inherits the same alignment of calendar and
diagnosis. (b)~The audit's five questions (\S\ref{sec:setup}) and this
benchmark's answers on one AUROC axis, chance at 0.5. Dark markers are the
acquisition date and green markers the thermogram; a filled marker is the answer
with its 95\% interval, a hollow marker the reference it is read against. Q1:
the date alone (the fitted arm of Table~\ref{tab:channels}; scored raw it reads
0.981, the statistic \S\ref{sec:controlled} uses) against the best imaging arm.
Q2: the acquisition year decoded from the pixels among benign patients. Q3: the
year carried by a diagnostic score among benign patients, for the one arm, of
the seven that pass the quality gate, that carries it (grey ticks: the others). Q4: the date plus the best
imaging arm against the date alone, $\Delta = +0.003$ [$-0.008$, $+0.015$].
Q5: the frontal probe on the date-overlap cohort against the full cohort. Each
number is read from the experiment that produced it.}
\label{fig:audit}
\end{figure}

\textbf{Notation.} $X$ is the image, $Y$ the patient-level diagnosis, $C$ a
candidate non-image channel. The audit asks five questions (Fig.~\ref{fig:audit}b):
Q1, confounder strength ($C \to Y$); Q2, legibility ($X \to C \mid Y$); Q3,
exploitation (dependence of the diagnostic score $s(X)$ on $C$ within $Y$); Q4,
incremental value ($\Delta = \mathrm{AUROC}(X{+}C) - \mathrm{AUROC}(C)$, paired); and
Q5, controlled discrimination ($X \to Y \mid C$), by matched cohorts and by a
conditional pairwise statistic. The arrows denote prediction tasks, not causal
effects. How a candidate $C$ should be read depends on context: the acquisition
date is an acquisition-related candidate, whereas age may be clinically informative.
Throughout, ``exploitation'' denotes the operational Q3 test of score--$C$
dependence within diagnosis strata; a positive test does not by itself identify a
causal mechanism.

\textbf{Terminology.} A \emph{patient visit} is everything collected from one
patient at one appointment (here a static series and a dynamic sequence); its
\emph{acquisition date} is the calendar date recorded for it; an
\emph{acquisition-date group} is the set of patients sharing one. An
\emph{acquisition session} proper, a contiguous collection under one
documented setup, is only proxied by a date group, since DMR-IR records the
date and not the setup: we therefore say ``date group'' for what the metadata
guarantees, and \emph{acquisition signature} for the room, operator and
camera-state mechanism a session would impose. For the same reason the protocol
below is \emph{patient-disjoint} rather than leakage-free: it guarantees that
no patient crosses a fold, and nothing more. ``Leakage-free'' in the title names
the field's usage of the term, which is the paper's subject.

\textbf{Primary cohort.} DMR-IR (Hugging Face mirror
\texttt{SemilleroCV/DMR-IR}): after excluding healthy volunteers (a
recruitment-channel shortcut worth AUROC 0.585 on its own) and collapsing
patients entered twice~\cite{perezmartin2023anomalies}, 236 patients (140
benign / 96 malignant), 6051 thermograms over five static view slots and a
dynamic sequence (medians 5 and 22 frames per patient): an \emph{all-frames}
arm reads all of them, a \emph{frontal} arm the one static anterior image.
Benign patients were acquired 2012--2015 and
malignant patients 2013--2020; the 236 patients occupy 86 acquisition dates, 83
of which contain a single class.

\textbf{Second cohort.} The published dataset of Rodr\'iguez-Guerrero et
al.~\cite{rodriguezguerrero2024breastthermography}: 119 patients (84 benign /
35 malignant), 357 thermograms, single centre and camera, 2020--2023
(Appendix~\ref{sec:cohort2}). Its dominant non-image
channel is patient age (AUROC 0.759), not the calendar; on DMR-IR age sits at
chance, reaching 0.495 [0.415, 0.576] scored directly.\footnote{The fitted age
arm of Table~\ref{tab:channels} reads 0.416, below chance: the signature of a
null under cross-validation, not an inverse association
(Appendix~\ref{sec:age}).} The two
cohorts therefore carry \emph{different} confounders, which is what makes the
second one an internal control on the audit itself.

\textbf{Protocol.} Five repetitions of stratified group five-fold
cross-validation with patient identity as the grouping variable; all frames of
a patient stay together; frame predictions are averaged within patient before
any interval is computed; 95\% CIs are patient-level bootstrap percentiles
(2000 resamples); paired comparisons use identical folds. Encoders
(ConvNeXt-T~\cite{liu2022convnext}, Swin-T~\cite{liu2021swin},
ResNet-50~\cite{he2016resnet}) and regimes (frozen probe,
fine-tuned) were pre-specified; both frontal ResNet-50 arms failed a pre-specified
quality gate (near-constant predictions in $\geq$6 of 25 folds) and are reported as
failed configurations, not results. A regression test re-checks every
prediction file against its declared folds. Code, configurations and aggregate
statistics are publicly available; the per-patient files are regenerated from the
public datasets (Appendix~\ref{sec:repro}).

\section{Results}
\label{sec:results}

\subsection{Q1, strength: the calendar outperforms the image}
Table~\ref{tab:channels} evaluates every channel on identical patient-level
folds. Acquisition date alone reaches 0.980 [0.967, 0.992]; the strongest
imaging arm, a Swin-T probe over all frames of a patient, reaches 0.949
[0.918, 0.973]; no imaging arm exceeds the calendar. This ordering does not
depend on the encoder: fine-tuning lifts the frontal view from 0.815 to
0.923, and the calendar is above every arm throughout (full encoder table in
Appendix~\ref{sec:encoders}; two runs of the frozen frontal configuration give 0.815 and
0.811, which bounds rerun variance here). Restricting to the calendar years
2013--2015, the years in which both classes were acquired, does not dissolve the
confounder: on those 114 patients the date still scores 0.914. (The narrower
date-overlap window of Fig.~\ref{fig:audit}a, from the first malignant to the last
benign acquisition, holds 60 patients and is the cohort of \S\ref{sec:controlled}.)

\begin{table}[t]
\centering
\small
\caption{Every information channel on DMR-IR, same patient-level folds,
patient-level bootstrap CIs. The image-level row reproduces the literature's
protocol and is contaminated by construction (0.9998 of test frames from seen
patients; 233/236 patients on both sides of the split).}
\label{tab:channels}
\begin{tabular}{lc}
\toprule
Channel & AUROC [95\% CI] \\
\midrule
Acquisition date alone & 0.980 [0.967, 0.992] \\
Tabular record, all fields & 0.952 [0.921, 0.976] \\
Clinical text, TF--IDF & 0.927 [0.888, 0.960] \\
\midrule
Thermogram, image-level split & 0.968 [0.945, 0.985] \\
Thermogram, all frames, Swin-T & 0.949 [0.918, 0.973] \\
Thermogram, all frames, ConvNeXt-T & 0.922 [0.882, 0.955] \\
Thermogram, frontal, fine-tuned & 0.923 [0.887, 0.954] \\
Thermogram, frontal, frozen probe & 0.815 [0.760, 0.867] \\
\midrule
Age alone, fitted & 0.416 [0.342, 0.498] \\
\bottomrule
\end{tabular}
\end{table}

\subsection{Q2--Q3, legibility and exploitation: in the pixels, and in
one arm's score}
\label{sec:legibility}
\begin{figure}[t]
\centering
\includegraphics[width=\linewidth]{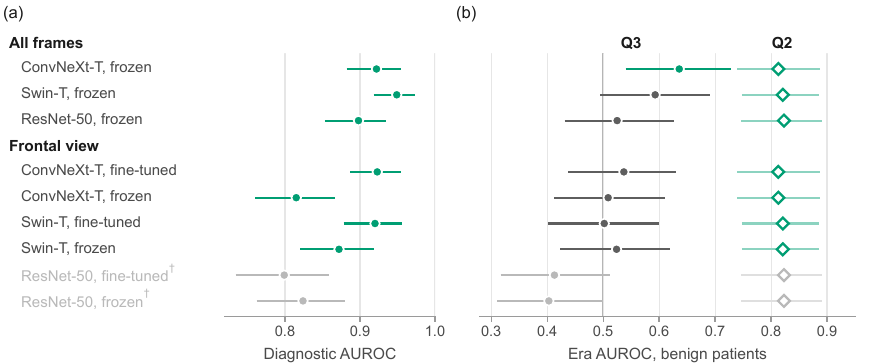}
\caption{Legibility against exploitation, one row per imaging arm, with 95\%
intervals; \emph{all frames} arms pool every frame of a visit, \emph{frontal}
arms read the one static anterior image. (a)~Diagnostic AUROC, malignant versus
benign. (b)~AUROC for the acquisition era (2012 versus 2013) among benign
patients, read two ways: from the image (Q2, open diamonds; decoded once per
backbone from its frozen representation, so arms sharing a backbone share it)
and from the arm's own diagnostic score (Q3, filled circles; green where the
permutation test is significant, $p = 0.005$). The vertical line marks chance.
Diagnostic AUROC varies by more than 0.11 across arms; the era is decodable at
0.81--0.82 in every one. Legibility is uniform; what varies is whether an arm's
own score carries the era, and only one does (Table~\ref{tab:exploitation}; all
runs in Appendix~\ref{sec:exploitation}). $\dagger$: configuration failing the
pre-specified quality gate.}
\label{fig:shortcut}
\end{figure}

A confounder in the metadata could be irrelevant to a model that never reads
metadata. It is not confined there. Trained to predict the acquisition era
\emph{among patients who share a diagnosis} (so the label cannot act as a
proxy), every encoder recovers the year: within the 133 benign patients (64
acquired in 2012, 69 in 2013) ConvNeXt-T reaches 0.813 [0.740, 0.887], and the
three backbones span 0.813--0.823 (Fig.~\ref{fig:shortcut}); a rerun of it
reads 0.814, bounding rerun variance (Appendix~\ref{sec:encoders}).

Acquisition-era information is recoverable from background pixels and from
summaries that discard spatial geometry. Twelve summary statistics of temperature
over the pixels the distributed masks exclude (the background alone) separate the
two acquisition years of these same benign patients at 0.938 [0.892, 0.976], above
every encoder; the breast region
alone reaches 0.806 [0.729, 0.875], and a 32-bin temperature histogram of the
whole frame, which retains no geometry whatsoever, reaches 0.899 [0.841, 0.951]
(Appendix~\ref{sec:signature}).
These results are consistent with an acquisition-related shift in the temperature
distribution, possibly a global offset; because the acquisition environment was not
recorded, they do not establish room temperature as its cause. The paired
representation probes of Appendix~\ref{sec:finetuneera} show small differences
between frozen and fine-tuned representations, with intervals crossing zero.

Legibility is not use, and the distinction turns out to matter more than we
expected. The direct test asks whether a \emph{diagnostic} model's scores
separate the eras where the diagnosis is held constant. Applied to all nine
imaging arms over ten runs, one configuration having been run twice and the two
frontal ResNet-50 arms carrying no inference since they failed the quality gate, with
patient-level bootstrap intervals and a permutation test
(Table~\ref{tab:exploitation}; full table in Appendix~\ref{sec:exploitation}), it
returns a much
narrower answer than legibility did: era information is demonstrable in the
scores of exactly one arm of the seven that pass the gate, the all-frames ConvNeXt-T
probe, at 0.636 [0.540, 0.729], $p=0.005$, a result that survives a Holm correction
across those seven configurations. The frozen frontal probe of that same encoder sits at chance
(0.509 [0.413, 0.610], $p=0.85$), and so do the two \emph{highest}-scoring
frontal arms, both fine-tuned: 0.923 AUROC with era at 0.537 ($p=0.47$), and
0.920 with era at 0.502 ($p=0.97$).

What separates the one arm that carries the shortcut is therefore not capacity.
It is how much of one visit the score aggregates: one static image against
every frame of it. Within
ConvNeXt-T, the paired contrast between the all-frames and frontal arms over the
133 benign patients both score is $+0.122$ [$+0.039$, $+0.202$] ($p=0.005$);
within Swin-T the same contrast is $+0.058$ [$-0.071$, $+0.178$] and does not
resolve.
Decodability, in short, does not imply exploitation: the confounder is equally
legible to every encoder, and one arm demonstrably carries it into its diagnostic
score. The single positive arm is nonetheless the informative one: averaging over
one visit's repeated observations sharpens an \emph{acquisition signature} rather
than a lesion, a reading to which \S\ref{sec:controlled} returns.

\begin{table}[t]
\centering
\small
\caption{Does a diagnostic model's score still separate the acquisition eras
among patients who share a diagnosis? Within the 2012/2013 benign stratum, with
patient-level bootstrap intervals and a two-sided permutation test on
$|\mathrm{AUROC}-0.5|$. Five arms shown; all ten runs in Appendix~\ref{sec:exploitation}.}
\label{tab:exploitation}
\begin{tabular}{lcc}
\toprule
Arm & Diagn. & Era within benign \\
 & AUROC & AUROC [95\% CI], $p$ \\
\midrule
ConvNeXt-T, all frames & 0.922 & 0.636 [0.540, 0.729], 0.005 \\
Swin-T, all frames & 0.949 & 0.593 [0.495, 0.691], 0.063 \\
ConvNeXt-T, frontal, fine-tuned & 0.923 & 0.537 [0.438, 0.630], 0.47 \\
Swin-T, frontal, fine-tuned & 0.920 & 0.502 [0.402, 0.599], 0.97 \\
ConvNeXt-T, frontal, frozen & 0.815 & 0.509 [0.413, 0.610], 0.85 \\
\bottomrule
\end{tabular}
\end{table}

\subsection{Q4, incremental value: no imaging arm improves on the calendar}
Paired on identical folds, no imaging arm improves on the date
(Table~\ref{tab:incremental}). The strongest arm moves the calendar from 0.980
to 0.984 under the label-free fixed combination, $\Delta = +0.003$
[$-0.008$, $+0.015$], $p=0.58$. Every weaker arm is worse than the date alone,
and so is the strongest arm's image score on its own ($-0.036$ [$-0.068$,
$-0.010$], $p=0.005$): the best image model in the paper is significantly
\emph{worse} than knowing the day of the appointment. We write ``no incremental
value was demonstrated'': the CI crosses zero, and no equivalence claim is made.

\begin{table}[t]
\centering
\small
\caption{Incremental value of the image given the date, paired on identical
folds. The fixed combination fits nothing to labels; the stacked variant is
exploratory. Improvement requires the CI to exclude zero from above; none
does. Frontal rows use the 231 patients holding a frontal view, all-frames rows
all 236; the \emph{alone} rows refit each arm's score on the same folds
(Appendix~\ref{sec:alone}).}
\label{tab:incremental}
\begin{tabular}{lcc}
\toprule
Arm & AUROC & $\Delta$ vs.\ date [95\% CI] \\
\midrule
Date alone & 0.980 & -- \\
\midrule
Frontal image alone & 0.811 & $-0.169$ [$-0.222$, $-0.120$] \\
Date + frontal (fixed) & 0.926 & $-0.054$ [$-0.084$, $-0.028$] \\
Date + frontal (stacked) & 0.977 & $-0.003$ [$-0.010$, $+0.004$] \\
\midrule
ConvNeXt-T all frames alone & 0.920 & $-0.060$ [$-0.098$, $-0.028$] \\
Date + ConvNeXt-T all fr.\ (fixed) & 0.967 & $-0.013$ [$-0.033$, $+0.003$] \\
Date + ConvNeXt-T all fr.\ (stacked) & 0.977 & $-0.003$ [$-0.015$, $+0.009$] \\
\midrule
Swin-T all frames alone & 0.945 & $-0.036$ [$-0.068$, $-0.010$] \\
Date + Swin-T all fr.\ (fixed) & 0.984 & $+0.003$ [$-0.008$, $+0.015$] \\
Date + Swin-T all fr.\ (stacked) & 0.984 & $+0.004$ [$-0.007$, $+0.014$] \\
\bottomrule
\end{tabular}
\end{table}

\subsection{Q5, controlled discrimination: what survives}
\label{sec:controlled}
\begin{figure}[t]
\centering
\includegraphics[width=\linewidth]{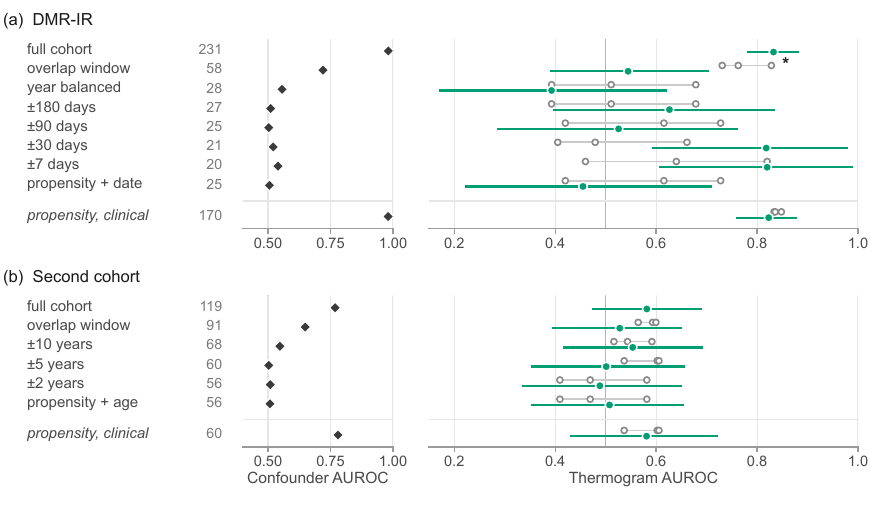}
\caption{The confounder-control ladder on (a)~DMR-IR, where the confounder is
the acquisition date, and (b)~the second cohort, where it is age; one row per
rung, with the number of patients scored. Left: the confounder's own AUROC once
the rung controls it (dark diamonds). Right: the thermogram's AUROC on that
cohort with its 95\% interval (green), beside three cohorts of the same class
sizes drawn with the confounding intact (open circles). The third element is
what makes the first two interpretable: a fall is attributed to the control only
when all three draws lie above the rung's whole interval ($*$), which only the
DMR-IR overlap rung achieves. The last row (italic) is the null
control, propensity matching on clinical covariates alone, which leaves the
confounder in place. Vertical lines mark chance. The DMR-IR panel is the frozen
frontal probe, so each count is the rung's patients holding a frontal view
(\S\ref{sec:controlled}).}
\label{fig:ladder}
\end{figure}

\begin{figure}[t]
\centering
\includegraphics[width=\linewidth]{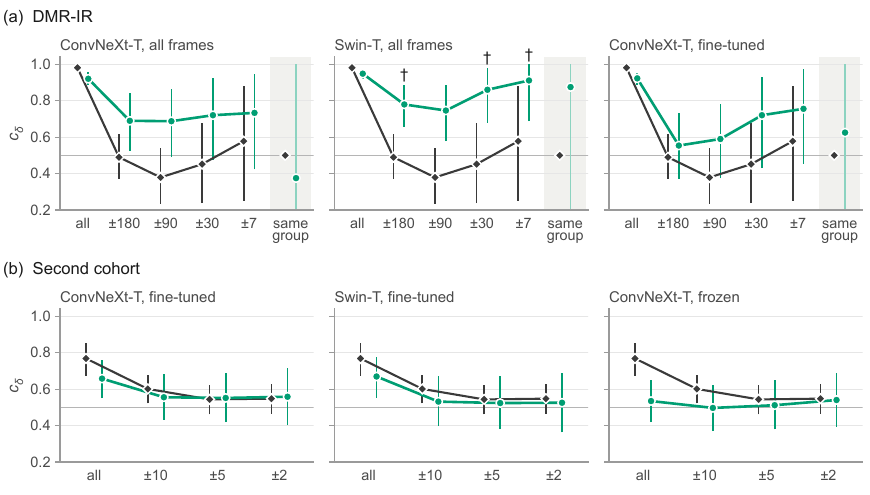}
\caption{Discrimination restricted to pairs the dominant confounder cannot
separate: the conditional AUROC $c_\delta$ against the caliper $\delta$ on
(a)~DMR-IR, $\delta$ in days of acquisition date, and (b)~the second cohort,
$\delta$ in years of age. One panel per imaging arm (green, with 95\%
intervals), each drawn against its control (dark): the confounder scored through
the identical statistic on the identical pairs, which must fall to chance
(horizontal line) for the imaging curve to be readable, and does. The calipers
leave 12\,878, 390, 256, 93, 45 and 8 admissible pairs in (a) and 2\,940,
1\,127, 678 and 335 in (b). \emph{Same group} (shaded) restricts to pairs sharing an
acquisition-date group, the closest the data come to holding acquisition constant
(\S\ref{sec:setup}): the eight such pairs DMR-IR contains, on which every imaging
interval spans the unit interval. $\dagger$: restricted cell surviving the Holm
correction over the pre-declared 25-cell family. The fine-tuned DMR-IR arm reads
the frontal view. Three of the family's five arms per cohort, by a priority
fixed in advance; the full grid is in Appendix~\ref{sec:conditional}.}
\label{fig:caliper}
\end{figure}

The clinical question is what survives once the calendar is controlled. Two
designs answer it, and they fail in different directions, which is why both are
reported (Figs.~\ref{fig:ladder},~\ref{fig:caliper}).

\textbf{Matched cohorts, with the controls that make them readable.} Restricting
to the 60 patients of the date-overlap window (Fig.~\ref{fig:audit}a), where the
date's own discrimination falls to 0.720, imaging discrimination falls to
0.544 [0.390, 0.705], from 0.833 on the full cohort. Every rung of the ladder,
including this reference, is the frozen frontal ConvNeXt-T probe fitted on a
fixed epoch budget without validation-based selection, so each rung scores its
cohort's patients holding a frontal view: 231 of 236 at the reference rung, 58
of 60 here. That recipe is also why its
full-cohort value is 0.833 rather than the 0.815 of Table~\ref{tab:channels},
whose frozen-probe arms restore the epoch of best validation
AUROC.\footnote{Epoch selection is active in the frozen-probe arms only; the
fine-tuned arms and every ladder rung run to a fixed budget, and rungs are
compared only against each other (Appendix~\ref{sec:selection}).}
The fall is attributable to the
control, not the shrinkage: three random 60-patient cohorts with the confounding
left intact reach 0.731--0.828, every one above the matched interval. The null
control behaves as a null control must: propensity matching on clinical
covariates alone leaves the date at 0.981 and discrimination at 0.823,
indistinguishable from its size controls (0.834--0.848). Balancing everything
\emph{except} the acquisition axis controls nothing. Stricter rungs (7--180-day
calipers, 20--28 patients) resolve nothing, since every matched estimate falls
inside its own control range, and are reported as descriptive only, in
Appendix~\ref{sec:ladder}; covariate balance for every rung is in
Appendix~\ref{sec:balance}.

\textbf{The same ladder on the all-frames arms.} A frontal ladder cannot speak
for arms that read every frame of a visit, so we refitted the overlap rung, its
three size controls and the null control on both all-frames configurations (all
236 and all 60 patients; full table in Appendix~\ref{sec:ladderall}). Two things
follow, in
different directions. Discrimination falls on
the overlap cohort for these arms too, from 0.950 to 0.757 [0.628, 0.878] for
Swin-T and from 0.922 to 0.788 [0.666, 0.898] for ConvNeXt-T, so the fall is not
an artefact of having tested a single view; and the null control replicates
exactly on both arms, with the date left at 0.981 and clinical-covariate
matching moving nothing (0.954 against controls 0.939--0.952; 0.916 against
0.902--0.934). But the attribution is not reproduced. Every size control again
lands above the matched estimate (0.853--0.894 and 0.807--0.871), yet at $n=60$
the matched interval overlaps them, so our pre-specified criterion, that every
control draw lie above the rung's whole interval, is met by the frontal arm and
by neither all-frames arm. For the all-frames arms we therefore claim only a
fall of the same sign and comparable size, with shrinkage not excluded.

That they fall \emph{less far} (to 0.76--0.79 rather than 0.54) is what
\S\ref{sec:legibility} would predict: the overlap rung weakens the confounder
rather than removing it (0.981 to 0.720), and the arms retaining most
discrimination there are the all-frames arms, the class of the one arm shown to
carry acquisition information into its scores (all-frames ConvNeXt-T; the era
statistic of all-frames Swin-T does not reach significance). The reading is
consistent, and the data cannot distinguish it from residual pathology signal.

\textbf{Conditioning the comparison instead of the cohort.} The second design
keeps every patient and recomputes the AUROC over only those benign/malignant
pairs whose acquisition dates differ by at most a caliper $\delta$: the conditional
AUROC $c_\delta$, in which ties count half and an unrestricted caliper recovers the
ordinary AUROC exactly. The estimator is self-checking: the date, scored through
the identical statistic, must fall to chance as the caliper narrows, and it does
(0.981 $\to$ 0.379--0.578). It retains more admissible comparisons than 1:1
matching in this analysis: 256 pairs at 90 days against 13 matched pairs. These
comparisons share patients, so the pair count is not an independent sample size
and does not by itself establish greater statistical power. Against that control
the arms diverge: the
frontal arms fall to 0.554--0.800 with intervals containing chance at every
caliper, while the Swin-T all-frames arm holds 0.779, 0.746, 0.860, 0.911 at
180, 90, 30, 7 days. Because that is a grid of five arms by five calipers, a
per-cell permutation test is Holm-adjusted over the 25-cell family declared in
advance; three of the twenty restricted cells survive, all of them Swin-T all
frames: 180 days ($p=0.023$), 30 days ($0.023$) and, most tellingly, 7 days
($0.008$). No cell of any other arm survives, including the ConvNeXt-T
all-frames arm of \S\ref{sec:legibility}. One arm survives conditioning at three
calipers of four; the rest of the grid is consistent with chance
(Appendix~\ref{sec:conditional}).

\textbf{Why the surviving signal is not yet thermography.} A date caliper
bounds elapsed time; it does not hold the acquisition \emph{session} constant.
Of the 45 pairs admitted at $\pm$7 days, only 8 share a date group, and 7 of
the 10 date groups those pairs draw on are label-pure. A model reading an
acquisition signature (room, operator, camera state) would order most
date-comparable pairs correctly while the date itself ordered none, which is the
pattern observed; and \S\ref{sec:legibility} showed that what puts acquisition
information into a diagnostic score is precisely the pooling of one visit's
frames, which is what these arms do. The arm that survives here is Swin-T all
frames, whose own era statistic (0.593) does not reach significance: the
mechanism is established on ConvNeXt-T, and we do not transfer it as though it
had been measured on Swin-T.

The comparison that would decide it is within date groups containing both
classes, and DMR-IR contains 8 such pairs in total: the Swin-T all-frames arm
orders 7 of 8 correctly, with an interval spanning the unit interval. Nothing can be
concluded from eight pairs, \emph{and that is the finding}. The benchmark
does not return a negative answer; it lacks the acquisition structure required
to return any answer at all.

\subsection{The audit on a second cohort: a legitimate confounder}
\label{sec:second}
On the second cohort the dominant
channel is age (0.759 [0.665, 0.844]; malignant patients average twelve years
older). No imaging arm exceeds it: the fine-tuned anterior arm reads 0.654
[0.543, 0.756], and the breast-level arm (fine-tuned ConvNeXt-T on the two oblique
views, one per breast, each patient scored by the higher of their two breasts), the
one paired against age, is worse by $-0.140$ [$-0.262$, $-0.018$] from its own 0.618. Adding the image to age
does not help ($-0.028$ [$-0.096$, $+0.037$], fixed). Under the same two controls the
pattern repeats:
matching on age within five years reduces age's own discrimination from 0.769
to 0.503 and the image reaches 0.501 [0.353, 0.657], while propensity matching
that retains the same 60 patients with age uncontrolled (0.780) reaches 0.581
[0.429, 0.723].\footnote{Ladder rungs and the conditional statistic score age
directly, with nothing fitted. Age therefore reads 0.769 on the full cohort and
0.780 on the 60 patients the propensity rung retains, against 0.759 fitted; the
fine-tuned encoder reads 0.654 and 0.658 in two runs of one configuration.}
Conditionally, both fine-tuned encoders fall from 0.658--0.669 unrestricted to
0.524--0.558 within ten-, five- and two-year calipers, every interval containing
chance, and 678 admissible pairs remain at five years. Because pairs share
patients, precision must be judged from the patient-level intervals rather than from
the pair count alone.

The exploitation test transfers too, and it is what makes the null above
legible. Splitting each diagnosis at its own median age, every arm that
discriminates at all separates older from younger patients \emph{within} a
diagnosis: on the 79 benign patients off their own median, the Spearman correlations
of the scores with untransformed age are $+0.29$ for fine-tuned ConvNeXt-T and $+0.35$
for Swin-T ($p\leq0.01$), and both survive a Holm correction across the three
configurations that pass the quality gate. The corresponding median-split AUROCs,
0.642 [0.513, 0.765] and 0.630 [0.504, 0.760], carry uncorrected permutation
$p$-values ($p=0.030$ and $p=0.048$) that do not survive the same correction; the frozen probe, whose
diagnostic discrimination is 0.535, sits at chance (0.501 [0.370, 0.630],
$p=0.99$; Appendix~\ref{sec:exploitation2}). Age is a legitimate clinical predictor, so
this is not the validity
failure the calendar is on DMR-IR, but it is the mechanism behind the
preceding paragraph: what these arms have that the frozen probe lacks is
consistent with an age-correlated appearance, and once age is held fixed there
is nothing left for them to separate. Two cohorts, two different dominant
channels, one audit, and the same outcome on every question except legibility
(Fig.~\ref{fig:caliper}b; Appendix~\ref{sec:exploitation2}).

\section{Discussion}
The audit separates four things that a single AUROC conflates. What does the
confounder predict? More than the image does. Does the image reveal it? Fluently,
in every encoder. Does the model use it? In one arm of the seven that pass the
quality gate, the one pooling every frame of a visit. What remains once it is controlled? Either nothing demonstrable,
or a residue the benchmark cannot attribute.

\textbf{Legibility is not exploitation.} The gap between the second and third
answers is the finding we most want carried away. A confounder can be legible to
every encoder, at 0.81--0.82 for all three here, and still be
demonstrably read by almost none of the arms built on them: one arm of the seven
that pass the quality gate, and not the strongest. Exploitation must therefore be measured on the diagnostic scores;
a decoder establishes only that a shortcut is \emph{available}, not that it was
taken. The converse matters as much. On the second cohort the same test is
positive for every arm that discriminates at all, and the channel there is
patient age, a legitimate clinical predictor. The test reports what an arm
tracks; whether that is a defect is a question about the channel, not about the
statistic, which is why the second cohort validates the method rather than
convicting a second dataset.

\textbf{Beyond thermography.} Site, scanner, era, device generation and care
pathway occupy the role the calendar occupies here, in radiology, pathology and
dermatology alike, and a benchmark result reported without these controls is a
measurement of an unidentified mixture. We do not claim that breast thermography
carries no signal; we claim, and show, that this benchmark cannot demonstrate
that it does.

\begin{figure}[t]
\centering
\includegraphics[width=\linewidth]{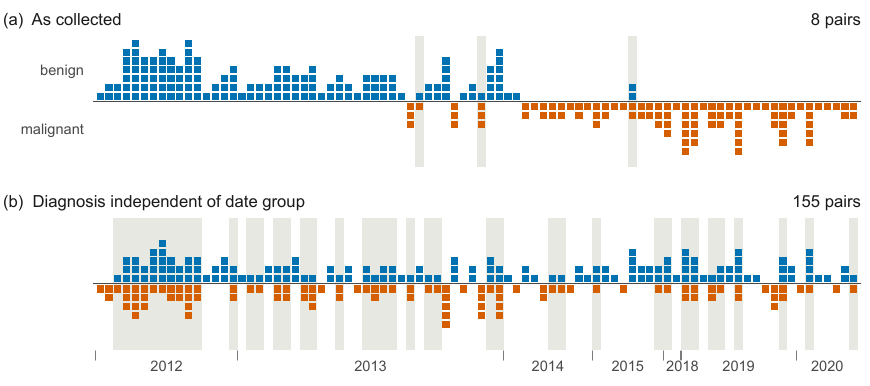}
\caption{\textbf{Cross-label acquisition support: what a patient-level split
cannot supply.} Each column is one acquisition-date group of DMR-IR, ordered by
date and marked by year below; each square is one patient, benign (blue) above
the line and malignant (orange) below. Shaded columns hold both classes, the
only groups that form a cross-label pair at fixed acquisition; the count of such
pairs is given on the right. (a)~As collected: 3 of the 86 groups hold both
classes, forming 1, 3 and 4 pairs. (b)~The same patients, groups and class
balance, with the diagnosis assigned independently of the group: 40 groups hold
both classes, and the draw shown is one whose pair count equals the median over
10\,000 permutations (95\% range 136--171). A patient-level split partitions
either panel equally well; only (b) holds the comparisons that decide whether a
model reads pathology or acquisition.}
\label{fig:support}
\end{figure}

\section{From audit to benchmark repair}
\label{sec:repair}
The audit reports what a benchmark is measuring. What follows from that report
depends on which of three repairs the benchmark needs, and the three are not
interchangeable: one applies to a finished split, one to a finished evaluation,
and one only at collection time.

\textbf{Repairing the split} fixes the unit of observation. Patient-level
grouping keeps one patient off both sides of a partition; where several images
share an acquisition occasion the unit is larger, since DMR-IR's static and
dynamic frames are one visit and not twenty-odd observations, so grouping
should follow the hierarchy
patient within session within site or device. That is worth doing, and it is not
what fails here. Grouping constrains which rows face each other across a fold and
cannot touch a population-level association between acquisition and diagnosis:
the date orders these 236 patients at 0.981 as a property of the cohort, and every
partition inherits that ordering intact. Nor does deleting the non-overlapping
years repair it (\S\ref{sec:results}). Split independence is a precondition, not
evidence.

\textbf{Repairing the evaluation} changes which comparisons the metric averages
over. The AUROC is $P[s(X^{+}) > s(X^{-})]$ over all benign/malignant pairs; the
statistic of \S\ref{sec:controlled} restricts it to admissible pairs,
$P[s(X^{+}) > s(X^{-}) \mid d(C^{+},C^{-}) \leq \delta]$, and recovers it as
$\delta$ grows. It needs no retraining and no new data, and it carries its own
falsification: the confounder, scored through the identical statistic on the
identical pair set, must fall to chance, or the restriction has controlled
nothing. It does on both cohorts. Where the admissible set stays large (256
pairs at $\pm$90 days here, 678 at $\pm$5 years on the second cohort) this
turns an uninterpretable aggregate into a statement about comparisons the
confounder cannot explain, on data already collected. But the exchange rate
between control and precision is fixed by the dataset, not by the analyst:
12\,878 admissible pairs unrestricted, then 390, 256, 93 and 45 as the caliper
narrows, and 8 when the pair must share an acquisition-date group.

\textbf{Repairing the dataset} is the only operation that moves that exchange
rate, and the only one that cannot be performed afterwards. What it supplies is
cross-label support in the acquisition conditions: for the conditions $c$ the
cohort spans, both $P(Y{=}0 \mid C{=}c) > 0$ and $P(Y{=}1 \mid C{=}c) > 0$, with
enough patients at each to estimate anything. This is an overlap condition, not
a theorem: where it fails, the data hold no comparison in which the label varies
and acquisition does not, and no statistic computed on them separates the two.

DMR-IR fails it almost entirely (Fig.~\ref{fig:support}). Three of its 86
date groups hold both classes; those three hold ten patients and yield eight of
the 13\,440 benign/malignant pairs the cohort forms. Had diagnosis been
independent of the calendar, the same 236 patients in the same 86 date groups
would have yielded a median of 155 [136, 171]; none of 10\,000 permutations fell
as low as eight (Appendix~\ref{sec:support}). The missing comparisons are a property of
how the cohort was
recruited, which is why \S\ref{sec:controlled} ends where it does.

\textbf{What can be mitigated, and what cannot be certified.} A split can be
recomputed and an evaluation restricted; support cannot be added to a finished dataset.
This is where confounding separates from the leakage it is routinely confused
with~\cite{kapoor2023leakage}: leakage is a defect of the protocol, and the same
patients repartitioned give an honest estimate, whereas confounding of this
severity is a defect of the sample. Mitigation is a different matter, and on DMR-IR it
is plausible. Because the signature is a global intensity statistic
(\S\ref{sec:legibility}), intensity standardisation per frame, histogram matching to a
reference~\cite{nyul2000standardization} or batch correction~\cite{johnson2007combat}
should remove much of it; adversarial removal of a named
nuisance~\cite{ganin2016dann}, invariant risk minimisation~\cite{arjovsky2019irm} and,
on cohorts larger than either of ours, generative counterfactuals~\cite{pawlowski2020deepscm}
can suppress what remains. The date also varies continuously within each diagnosis,
which gives such methods within-class variation to work with. We have not run these
mitigations here, and we expect them to lower legibility. What none of them can supply is
the evidence that the suppression left pathology behind. That evidence comes from
comparisons in which the label varies at fixed acquisition, and this benchmark holds
eight. No method evaluated only on the confounded benchmark can certify that what it
retained is disease rather than a second acquisition correlate it was never told to
remove. Re-running the audit on mitigated data is therefore the right test of a
mitigation, and its last question is the one the benchmark cannot answer.

\textbf{Three properties, not one.} A benchmark supporting a claim about disease
information needs three things, of which the one the field checks is the
weakest: independent observational units across splits; cross-label support
across the acquisition conditions the cohort spans; and evaluation conditional
on the dominant candidate confounders, the confounder scored through the same
statistic as its control. Only the second is expensive, and only at collection
time: interleaving recruitment so each era, site and device contributes both
classes costs no more than scheduling. The third costs a script.

One corollary is worth stating because it is tempting: withholding the
acquisition metadata is not a repair. On DMR-IR the calendar survives the
removal of the date field, since two age columns of the released record
reconstruct elapsed time at Pearson 0.990 (Appendix~\ref{sec:metadata}). Omitting the
acquisition identifiers costs the ability to measure a confounder, not the confounder.

\section{Limitations}
Both cohorts are small (96 and 35 events); unpaired differences below roughly 0.10
AUROC are not resolvable (paired differences on identical folds resolve smaller ones,
as in Table~\ref{tab:incremental}), every ``no improvement'' means \emph{not demonstrated}
rather than excluded, and matched-versus-control comparisons are unpaired by
construction. On the all-frames arms the matched interval at $n=60$ overlaps the
size controls, so shrinkage is excluded only for the frontal arm. Exploitation
is demonstrated on one arm of the seven that pass the quality gate and one encoder
of three, and the Swin-T
all-frames arm, whose conditional discrimination survives
\S\ref{sec:controlled}, does not itself reach significance on the era
statistic (0.593, $p=0.063$): the acquisition-signature reading of that survival
rests on mechanism and counting, not on a measurement of that arm. The acquisition
mechanism itself is observed only indirectly: DMR-IR records the date, not the
session, room, operator or camera state. On the second
cohort legibility returns a null (0.476 [0.344, 0.608] within benign, 0.602
[0.394, 0.804] within malignant) on a weak test: splitting each diagnosis at its
own median age leaves 79 benign and 34 malignant patients. That cohort
establishes that its arms carry age, not that age is legible in it, and because age is a
legitimate predictor it validates the audit's mechanics rather than the specific
benchmark-failure reading established on DMR-IR. Both cohorts are thermographic: the
audit is stated for any modality in which acquisition varies with the label, but its
behaviour on site, scanner or device confounders in other modalities, its calibration
against confounders of known strength, and the effect of the mitigations discussed in
\S\ref{sec:repair} are not tested here. The encoder set is pre-specified and small by
design, and one
of three backbones failed its quality gate in both regimes. Zero-shot
transfer to the second cohort reaches chance but confounds domain shift with
absent signal, and is reported only in Appendix~\ref{sec:transfer}, where it is not
load-bearing.

\section{Conclusion}
A strict patient-disjoint protocol certified a 0.95-AUROC model on a widely used
benchmark of an imaging subfield, yet the acquisition date alone scores higher,
the model adds nothing to the date on paired folds, and what it retains once the
date is controlled rests on comparisons the benchmark cannot attribute. Splitting by
patient removes duplicated patients; it does not remove the confounder, and
on this benchmark the confounder is stronger than the image, visible in the
image, carried into the score of the arm that pools a visit's frames, and, at
the decisive comparison, uncontrollable with the data that exist. Separating what that
establishes:
\emph{demonstrated}, that the confounder outpredicts the image, is legible in
it, is carried into one arm's score, and takes the frontal arm's discrimination
with it when controlled; \emph{supported but not established}, the same collapse
on the all-frames arms and the acquisition-signature reading of what survives
conditioning; \emph{unresolved here}, whether thermography carries disease
information at all: eight cross-label same-date pairs answer that in neither
direction. A benchmark
needs more than independent patients: it needs enough cross-label acquisition
support to separate pathology from acquisition. Mitigation can suppress a confounder
after the fact, but that support is bought when the data are collected.
Leakage-free is not confounder-free.

\bibliographystyle{unsrtnat}
\bibliography{references}

\clearpage
\appendix
\FloatBarrier

\section{Reproducibility and reading conventions}
\label{sec:repro}
The analysis code, the experiment configurations and the aggregate statistics from
which every number in this paper is computed are available at
\url{https://github.com/ignaziogallo/breast_thermography_reliable_ai}, with a single
\texttt{README.md} describing dataset preparation and execution. Every
\texttt{scripts/\dots} and \texttt{results/\dots} path cited below refers to that
repository. Neither the datasets nor any per-patient file derived from them (manifests,
folds, predictions) is redistributed, because each carries a patient identifier, a
diagnosis and often an acquisition date and an age: the README states which public
releases are required, and the repository rebuilds the manifests and folds from them on
a CPU and the predictions on a GPU. Every run records the SHA-256 of the manifest and
folds it consumed, so a rebuilt cohort can be checked against the one that was run. The
initial weights every fold starts from (the pretrained encoder and its seeded classifier
head) are distributed with the code; with weights downloaded afresh, the frozen probes
move by about 0.01 AUROC, within the reported intervals. The analyses in these
appendices read the predictions and run on a CPU in minutes; a GPU is needed only to
regenerate them.

Every numeric literal in the paper is checked against the result tree by
\path{scripts/36_consistency_sweep.py}, which maps each claim to the file, row and
column that produced it and fails when one drifts.

\textbf{Reading conventions.} Unless a caption says otherwise: AUROC is patient-level, with
frame predictions averaged within patient before any interval is computed; intervals are
95\% patient-level bootstrap percentiles over 2000 resamples; $p$-values are two-sided
permutation tests over 10\,000 relabellings; $n$ counts patients; paired quantities are
computed on identical folds; and $\dagger$ marks a configuration that fails the
pre-specified quality gate, which allows at most 5 of 25 folds to emit predictions with
standard deviation below 0.05. Configurations failing the pre-specified quality gate are
reported for transparency but are excluded from the results and from all inferential
claims. As in the main text, the protocol is called \emph{patient-disjoint} rather
than leakage-free, an \emph{acquisition-date group} is the set of patients sharing one
recorded date, and \emph{acquisition signature} names the room, operator and camera-state
mechanism a date group only proxies.

\FloatBarrier
\section{Encoder robustness: the full encoder table}
\label{sec:encoders}
Pre-specified set: three backbones, two regimes, identical folds. Both frontal
ResNet-50 arms failed the quality gate, collapsing on 20 and 25 folds against the 5
it allows.
The era column is the same backbone trained to recover acquisition era among
benign patients only; values repeat across regimes because era decoding was
performed once per backbone (frozen representation), in the
\texttt{experiment\_dmr\_12\_epoch\_*} runs, which are the family every era
number in the paper is read from. An earlier, independent run
of the identical ConvNeXt-T configuration
(\texttt{experiment\_dmr\_03\_epoch\_within\_benign}: same manifest, same folds,
same seed) reads 0.814 [0.739, 0.886] against the 0.813 [0.740, 0.887] of
Table~\ref{tab:encoders}, which is the rerun variance of this decoder and the only reason
both numbers exist. Appendix~\ref{sec:finetuneera} shows that this column understates the
era information a representation holds, and that fine-tuning does not change it.

\begin{table}[h]
\centering
\small
\caption{Diagnostic and acquisition-era discrimination for every encoder and regime on
DMR-IR, on identical patient-level folds. Diagnosis AUROC is scored on the arm's own
patients (231 for the frontal rows, 236 for the all-frames rows); era AUROC is scored
within the 133 benign patients holding a frontal view. $\dagger$: configuration failing the
pre-specified quality gate.}
\label{tab:encoders}
\begin{tabular}{lccc}
\toprule
Encoder / regime & Frames & Diagnosis AUROC & Era AUROC \\
\midrule
ConvNeXt-T, frozen & frontal & 0.811 [0.755, 0.864] & 0.813 [0.740, 0.887] \\
ConvNeXt-T, fine-tuned & frontal & 0.923 [0.887, 0.954] & 0.813 [0.740, 0.887] \\
Swin-T, frozen & frontal & 0.872 [0.820, 0.918] & 0.821 [0.748, 0.885] \\
Swin-T, fine-tuned & frontal & 0.920 [0.878, 0.956] & 0.821 [0.748, 0.885] \\
ResNet-50, frozen$^{\dagger}$ & frontal & 0.824 [0.762, 0.880] & 0.823 [0.747, 0.891] \\
ResNet-50, fine-tuned$^{\dagger}$ & frontal & 0.799 [0.734, 0.858] & 0.823 [0.747, 0.891] \\
ConvNeXt-T, frozen & all frames & 0.922 [0.882, 0.955] & 0.813 [0.740, 0.887] \\
Swin-T, frozen & all frames & 0.949 [0.918, 0.973] & 0.821 [0.748, 0.885] \\
ResNet-50, frozen & all frames & 0.898 [0.853, 0.935] & 0.823 [0.747, 0.891] \\
\bottomrule
\end{tabular}
\end{table}

\FloatBarrier
\section{Exploitation, all arms}
\label{sec:exploitation}
Table~\ref{tab:exploitation} reports five rows; Table~\ref{tab:exploit_all} is the
complete set: ten runs
of nine distinct imaging arms, the two ConvNeXt-T frontal frozen rows being one
configuration trained twice. The statistic is the
AUROC with which an arm's own malignancy scores separate the 2012 and 2013
acquisition eras \emph{among benign patients only}, so the diagnosis cannot act
as a proxy. Reproduced by \path{scripts/34_exploitation_interval.py}.

Only the top row is distinguishable from chance. Across the ten runs,
Spearman(diagnostic AUROC, era AUROC) $=0.73$ ($p=0.016$), but it falls to
$0.67$ ($p=0.071$) once the two arms failing the quality gate are dropped, so we do not
report it as evidence. Those two arms separate the eras in
\emph{reverse}, which is one more reason not to read them as results.

\begin{table}[h]
\centering\small
\caption{Acquisition-era (2012 versus 2013) discrimination from each diagnostic arm's
malignancy score among benign patients. Each arm is scored on the benign patients it
covers: 136 for the all-frames rows, 133 for the frontal rows, which are restricted to the
231 patients holding a frontal view. An all-frames row pools every frame of a patient
visit, static series and dynamic sequence alike (\texttt{protocol=all}); a frontal row
reads the one static anterior image. $\dagger$: configuration failing the pre-specified
quality gate. $\ddagger$: a second,
independent run of the row below it, kept to show run-to-run variation (0.004 diagnostic,
0.003 era).}
\label{tab:exploit_all}
\begin{tabular}{llcc}
\toprule
Arm & Frames & Diagn. & Era within benign \\
 & & AUROC & AUROC [95\% CI], $p$ \\
\midrule
ConvNeXt-T, frozen & all fr. & 0.922 & 0.636 [0.540, 0.729], 0.005 \\
Swin-T, frozen & all fr. & 0.949 & 0.593 [0.495, 0.691], 0.063 \\
ConvNeXt-T, fine-tuned & frontal & 0.923 & 0.537 [0.438, 0.630], 0.47 \\
ResNet-50, frozen & all fr. & 0.898 & 0.525 [0.431, 0.627], 0.62 \\
Swin-T, frozen & frontal & 0.872 & 0.524 [0.423, 0.619], 0.62 \\
ConvNeXt-T, frozen$^{\ddagger}$ & frontal & 0.811 & 0.512 [0.416, 0.614], 0.82 \\
ConvNeXt-T, frozen & frontal & 0.815 & 0.509 [0.413, 0.610], 0.85 \\
Swin-T, fine-tuned & frontal & 0.920 & 0.502 [0.402, 0.599], 0.97 \\
ResNet-50, fine-tuned$^{\dagger}$ & frontal & 0.799 & 0.413 [0.317, 0.512], 0.080 \\
ResNet-50, frozen$^{\dagger}$ & frontal & 0.824 & 0.403 [0.311, 0.500], 0.053 \\
\bottomrule
\end{tabular}
\end{table}

Table~\ref{tab:exploit_contrast} isolates what separates the one positive arm from the
rest. Aggregating every frame of one patient visit, rather than reading one static view of
it, raises the acquisition information in the diagnostic score for ConvNeXt-T; for Swin-T
the contrast does not resolve. The ResNet-50 contrast, whose configuration fails the
quality gate, is significant but moves
from \emph{below} chance to chance, so in absolute terms its era information falls, and we
do not count it as support for the mechanism.

\begin{table}[h]
\centering\small
\caption{Era AUROC within benign patients, frontal versus all frames, same encoder and the
same 133 benign patients, so the difference is paired. The all-frames column is the
restriction of each all-frames arm to the benign patients that also hold a frontal view:
it reads 0.631, 0.582 and 0.520 here against the 0.636, 0.593 and 0.525 of
Table~\ref{tab:exploit_all}, which scores those same arms on all 136 benign patients they
cover. The contrast is computed on this restricted column. $\dagger$: configuration failing
the pre-specified quality gate.}
\label{tab:exploit_contrast}
\begin{tabular}{lccc}
\toprule
Encoder & Frontal & All frames & Paired $\Delta$ [95\% CI] \\
\midrule
ConvNeXt-T & 0.509 & 0.631 & $+0.122$ [$+0.039$, $+0.202$] \\
Swin-T & 0.524 & 0.582 & $+0.058$ [$-0.071$, $+0.178$] \\
ResNet-50$^{\dagger}$ & 0.403 & 0.520 & $+0.116$ [$+0.043$, $+0.194$] \\
\bottomrule
\end{tabular}
\end{table}

\FloatBarrier
\section{\texorpdfstring{Split protocol $\times$ unit of analysis}{Split protocol x unit of analysis}}
\label{sec:splitunit}
\begin{figure}[h]
\centering
\includegraphics[width=0.6\linewidth]{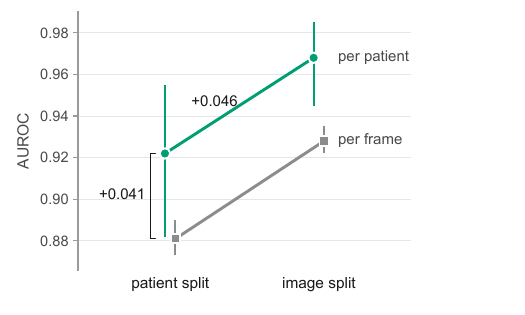}
\caption{The cells of Table~\ref{tab:splitunit} as intervals: ConvNeXt-T, split by
patient (contamination 0.0000) or by image (0.9998), scored per patient (green) or per
frame (grey). The rise of each line is the split effect ($+0.046$ scored per patient)
and the bracketed gap between the lines the unit effect ($+0.041$ under the
patient-level split). Both inflations are positive and of comparable size, and the
per-frame cells carry visibly narrower intervals because they count correlated frames
as independent observations.}
\label{fig:splitunit}
\end{figure}

Encoder, hyperparameters, epochs and folds identical across cells; only the
definition of a fold changes. Contamination is measured, not asserted. The split effect is
$+0.046$ and the unit effect $+0.041$; they compound to $\approx 0.09$ over the
patient-disjoint per-frame estimate. Table~\ref{tab:splitunit} and
Figure~\ref{fig:splitunit} show the same four cells.

\begin{table}[h]
\centering
\small
\caption{The four cells of split unit $\times$ analysis unit on DMR-IR, ConvNeXt-T.
\emph{Contamination} is the fraction of test frames whose patient also appears in the train
or validation partition of the same fold. The per-frame rows resample frames rather than
patients, deliberately: they reproduce how image-level work reports its intervals, and
are the one exception to the patient-level convention of Appendix~\ref{sec:repro}. Their
narrower intervals are optimistic for that reason.}
\label{tab:splitunit}
\begin{tabular}{lccc}
\toprule
Split unit & Analysis unit & AUROC & Contamination \\
\midrule
Patient & per patient & 0.922 [0.882, 0.955] & 0.0000 \\
Patient & per frame & 0.881 [0.873, 0.890] & 0.0000 \\
Image & per patient & 0.968 [0.945, 0.985] & 0.9998 \\
Image & per frame & 0.928 [0.922, 0.935] & 0.9998 \\
\bottomrule
\end{tabular}
\end{table}

\FloatBarrier
\section{\texorpdfstring{The \emph{alone} rows of the incremental table}{The alone rows of the incremental table}}
\label{sec:alone}
Table~\ref{tab:incremental} scores each arm twice, once alone and
once combined with the date, and both readings are computed from the arm's
stored prediction file by one estimator: the incremental pipeline
(\path{scripts/16_incremental_value.py}) refits the arm's patient score as
a single-feature logistic model on the same patient-level folds, so that the
\emph{alone} row and the two combination rows of a block differ only in which
features enter, never in how they were fitted. Refitting one feature cannot
reorder patients inside a fold, but it recalibrates each fold separately, and
the reported estimate pools the repeat-averaged probabilities across folds; the
pooled ordering therefore shifts slightly. The shift is small and, on all three
arms, downward (Table~\ref{tab:alone}).

\begin{table}[h]
\centering\small
\caption{The same stored prediction files read two ways, on DMR-IR. The patient set is not
the source of the gap: it is recorded as \texttt{n\_patients\_scored} in each run's
\texttt{config.json} and is identical down each row.}
\label{tab:alone}
\begin{tabular}{lccc}
\toprule
Arm & Patients & Table~\ref{tab:channels} & \emph{Alone} row \\
\midrule
Frontal, ConvNeXt-T frozen & 231 & 0.815 & 0.811 \\
All frames, ConvNeXt-T & 236 & 0.922 & 0.920 \\
All frames, Swin-T & 236 & 0.949 & 0.945 \\
\bottomrule
\end{tabular}
\end{table}

The frontal arms are restricted to the 231 patients holding a frontal view because that is
the set their own run scores; the all-frames arms score all 236. A third reading of the
same two all-frames files appears in Table~\ref{tab:conditional} at 0.920 and 0.948, where
every arm is cut to the 231 patients all five arms of that family cover.

One coincidence is worth flagging so it is not read as a discrepancy. The
frozen frontal probe's \emph{alone} row reads 0.811, and so does the
independent second run of that same configuration in Appendix~\ref{sec:encoders}. They
are different quantities that agree to three decimals: one is a refit of the 0.815 run, the
other a separate training run.

\FloatBarrier
\section{Full matched ladder (DMR-IR)}
\label{sec:ladder}
Every rung of Table~\ref{tab:ladder}, including the \texttt{original} reference, is the
frozen frontal ConvNeXt-T linear probe fitted on a fixed epoch budget with \emph{no}
validation-based model selection. Rungs are comparable to each other and to their own size
controls, and are not comparable to Tables~\ref{tab:channels}--\ref{tab:incremental},
which restore the epoch of best validation AUROC in their frozen-probe arms; that epoch
rule, and not the cohort, is why the reference rung reads 0.833 where
Table~\ref{tab:channels} reads 0.815 (Appendix~\ref{sec:selection}).

Only the \texttt{overlap} rung resolves anything. The stricter rungs (7--180-day calipers,
\texttt{year\_balanced}, \texttt{propensity\_w\_acq}) retain 20--28 patients, and every one
of their matched estimates falls inside its own size-control range, so they are reported as
descriptive only. The \texttt{propensity\_clinical} rung is the null control: matching on
clinical covariates alone leaves the confounder at 0.981 and moves discrimination nowhere.

\begin{table}[h]
\centering\small
\caption{The matched-cohort ladder on DMR-IR, every rung scored by the same frozen
frontal ConvNeXt-T probe. \emph{Confounder} is the acquisition
date scored raw on that rung's patients. $n$ is the matched cohort as the design defines it
on the patient manifest; \emph{scored} is how many of those the probe could read, since it
needs a static frontal view and five of DMR-IR's 236 patients hold none. \emph{Size
controls} is the range over three random cohorts of the same class sizes drawn with the
confounding left intact; \emph{above CI} records whether every one of them lies above the
rung's whole interval, the pre-specified attribution criterion. Runs:
\texttt{experiment\_dmr\_10\_matched\_*}.}
\label{tab:ladder}
\setlength{\tabcolsep}{4pt}
\begin{tabular}{lcccccc}
\toprule
Cohort & $n$ & scored & Confounder & Image AUROC & Size controls & Above CI \\
\midrule
original & 236 & 231 & 0.981 & 0.833 [0.780, 0.882] & -- & -- \\
overlap & 60 & 58 & 0.720 & 0.544 [0.390, 0.705] & 0.731--0.828 & yes \\
year\_balanced & 28 & 28 & 0.556 & 0.393 [0.169, 0.623] & 0.393--0.679 & no \\
days\_matched\_7 & 20 & 20 & 0.540 & 0.820 [0.606, 0.990] & 0.460--0.820 & no \\
days\_matched\_30 & 22 & 21 & 0.521 & 0.818 [0.591, 0.981] & 0.405--0.661 & no \\
days\_matched\_90 & 26 & 25 & 0.503 & 0.526 [0.286, 0.763] & 0.420--0.728 & no \\
days\_matched\_180 & 28 & 27 & 0.510 & 0.626 [0.396, 0.836] & 0.393--0.679 & no \\
propensity\_clinical & 174 & 170 & 0.981 & 0.823 [0.757, 0.879] & 0.834--0.848 & no \\
propensity\_w\_acq & 26 & 25 & 0.506 & 0.455 [0.221, 0.712] & 0.420--0.728 & no \\
\bottomrule
\end{tabular}
\end{table}

The gap between $n$ and \emph{scored} is where the 236 of the cohort description and the
231 of the frontal arms separate. It is also why the three size-control draws for the
\texttt{overlap} rung score 60, 59 and 57 against its 58: the design matches cohort size,
and the frontal view then removes a patient or two on either side.

\FloatBarrier
\section{The matched ladder on the all-frames arms}
\label{sec:ladderall}
The rungs that carry the inferential content, namely the date-overlap cohort,
its three size controls, and the propensity null control with its own three,
refitted on the two all-frames configurations (Table~\ref{tab:ladder_allframes}).
Identical cohorts, identical folds, identical frozen-probe recipe and fixed 15-epoch
budget; only the arm changes. These arms read every frame of a visit, so unlike the frontal
ladder of Table~\ref{tab:ladder} they score every patient of each cohort: 236, 60 and 174
respectively.
All 18 runs passed the prediction-spread check and patient-level contamination is zero
throughout. The \texttt{original} rows are refits of the two all-frames configurations
under this recipe, which is why they read 0.922 and 0.950 against the 0.922 and 0.949
of Table~\ref{tab:channels}: same configuration, separate run.

Two things follow, in different directions. Discrimination falls on the overlap cohort for
these arms too, so the fall is not an artefact of having tested a single view; and the null
control replicates exactly on both arms. But the attribution is not reproduced: at $n=60$
the matched interval overlaps the size controls, so for the all-frames arms we claim only a
fall of the same sign and comparable size, with shrinkage not excluded.

\begin{table}[h]
\centering\small
\caption{The overlap rung and the propensity null control refitted on the two all-frames
arms, with the frontal ladder repeated for comparison. \emph{Conf.}\ is the acquisition
date scored raw on that cohort; \emph{size controls} is the range over three
confounding-intact draws of the same class sizes. $*$ marks the only rung meeting the
pre-specified attribution criterion, every size-control draw above the \emph{whole} matched
interval.}
\label{tab:ladder_allframes}
\begin{tabular}{llccc}
\toprule
Arm & Cohort & Conf. & Image AUROC [95\% CI] & Size controls \\
\midrule
Frontal ConvNeXt-T & original & 0.981 & 0.833 [0.780, 0.882] & -- \\
& overlap & 0.720 & 0.544 [0.390, 0.705] & 0.731--0.828$^{*}$ \\
& null ctrl & 0.981 & 0.823 [0.757, 0.879] & 0.834--0.848 \\
\midrule
All frames Swin-T & original & 0.981 & 0.950 [0.920, 0.974] & -- \\
& overlap & 0.720 & 0.757 [0.628, 0.878] & 0.853--0.894 \\
& null ctrl & 0.981 & 0.954 [0.922, 0.977] & 0.939--0.952 \\
\midrule
All frames ConvNeXt-T & original & 0.981 & 0.922 [0.883, 0.956] & -- \\
& overlap & 0.720 & 0.788 [0.666, 0.898] & 0.807--0.871 \\
& null ctrl & 0.981 & 0.916 [0.869, 0.955] & 0.902--0.934 \\
\bottomrule
\end{tabular}
\end{table}

On the all-frames arms each size control still exceeds the matched point estimate, but not
the whole interval. The null control, propensity matching on clinical covariates, which
leaves the date at 0.981, moves nothing on any of the three arms, and that is what licenses
reading the overlap rung as being about acquisition rather than about matching per se.

\FloatBarrier
\section{Conditional discrimination, full grid (DMR-IR)}
\label{sec:conditional}
Table~\ref{tab:conditional} reports the full five-arm conditional-discrimination grid;
Figure~\ref{fig:caliper} visualises three of these five imaging arms.
The date control is excluded from the correction family: it is required to fail, and its
failure is what licenses reading the imaging rows. Of the 20 restricted cells only three
survive the correction, all in the all-frames Swin-T row: $\pm$180\,d
($p_{\mathrm{Holm}} = 0.023$), $\pm$30\,d ($0.023$) and $\pm$7\,d ($0.008$); its
$\pm$90\,d cell does not ($0.194$).

\begin{table}[h]
\centering\small
\caption{The conditional AUROC $c_\delta$, restricted to benign/malignant patient pairs
whose acquisition dates differ by at most the column's caliper, on the 231 patients all
five imaging arms cover. \emph{Same group} restricts to pairs sharing an acquisition-date group,
the closest the data come to holding acquisition constant (a shared date need not be a
shared session). $*$: bootstrap interval excludes 0.5,
uncorrected. $\dagger$: additionally survives a Holm correction at 0.05 over the
pre-declared family of 25 cells (5 imaging arms $\times$ 5 calipers), each cell tested by
permuting the patient scores with labels and dates, and therefore the admissible pair set,
held fixed (\texttt{scripts/35\_caliper\_multiplicity.py}). The \emph{same group} column is
not part of that family.}
\label{tab:conditional}
\setlength{\tabcolsep}{4pt}
\begin{tabular}{lcccccc}
\toprule
Score & All & $\pm$180d & $\pm$90d & $\pm$30d & $\pm$7d & Same group \\
\midrule
Date (control) & 0.981$^{*}$ & 0.490 & 0.379 & 0.452 & 0.578 & 0.500 \\
All frames, ConvNeXt-T & 0.920$^{\dagger}$ & 0.690$^{*}$ & 0.688 & 0.720 & 0.733 & 0.375 \\
All frames, Swin-T & 0.948$^{\dagger}$ & 0.779$^{\dagger}$ & 0.746$^{*}$ & 0.860$^{\dagger}$ & 0.911$^{\dagger}$ & 0.875 \\
Frontal, ConvNeXt-T, fine-tuned & 0.923$^{\dagger}$ & 0.554 & 0.590 & 0.720 & 0.756 & 0.625 \\
Frontal, Swin-T, fine-tuned & 0.920$^{\dagger}$ & 0.618 & 0.652 & 0.763 & 0.756 & 0.750 \\
Frontal, Swin-T, frozen & 0.872$^{\dagger}$ & 0.605 & 0.629 & 0.742 & 0.800 & 0.875 \\
\midrule
Admissible pairs & 12\,878 & 390 & 256 & 93 & 45 & 8 \\
\bottomrule
\end{tabular}
\end{table}

The two all-frames arms here are frozen probes. Every arm is scored on the 231 patients all
five cover, which is why the unrestricted column reads 0.920 and 0.948 for the all-frames
arms where Table~\ref{tab:encoders} reads 0.922 and 0.949 on all 236. The \emph{same group}
column rests on eight pairs, which is why it is kept out of the Holm family: with eight
pairs every interval spans the unit interval and no cell of it could be interpreted.

\FloatBarrier
\section{Cross-label acquisition support}
\label{sec:support}
The last column of Table~\ref{tab:conditional} is the only cell in which the recorded
acquisition date is held fixed, and it holds eight pairs. To determine whether the eight same-date
cross-label pairs result merely from cohort size or from the recruitment structure, we
compare the observed support with a permutation null.
\path{scripts/41_cross_label_support.py} counts the same two
quantities under the null in which the diagnosis is independent of the date
group: the same 236 patients, the same 86 date groups with the same sizes, labels
permuted. It reads the patient manifest only, with no pixels, no model and no
fold, and is deterministic given its seed.

No permutation reached either observed value (Table~\ref{tab:support}); the three date
groups that do hold both classes hold ten patients between them, six malignant and four
benign, which is where the eight pairs come from. The permutation is a description of the
recruitment design and not a test about any model: it says how far the collinearity of
label and date group is from what the group sizes alone would produce.
The draw shown in Figure~\ref{fig:support}b is the first draw of the same script whose
pair count equals the median.

\begin{table}[h]
\centering\small
\caption{Cross-label support in DMR-IR against 10\,000 permutations of the diagnosis with
the acquisition-date group sizes held fixed.}
\label{tab:support}
\begin{tabular}{lcc}
\toprule
Statistic & Observed & Null median [2.5, 97.5] \\
\midrule
Date groups holding both classes & 3 & 44 [38, 50] \\
Pairs sharing a date group & 8 & 155 [136, 171] \\
\bottomrule
\end{tabular}
\end{table}

\FloatBarrier
\section{Attribution of the metadata channel}
\label{sec:metadata}
Table~\ref{tab:metadata} partitions the tabular record into pre-diagnostic and
care-pathway fields and ablates clauses of the distributed text prompt.
Removing the care-pathway clause costs only 0.019, so the published
metadata-only result is not mainly pathway leakage. Separately, the released
tabular file's two age columns (\texttt{age\_at\_visit}, \texttt{age\_current};
0.500 and 0.377 alone) reconstruct the acquisition calendar through their
difference (Pearson 0.990 with true elapsed time; 0.946 together): any study
feeding every column to a tabular model inherits the temporal confounder
without touching a date field.

\begin{table}[h]
\centering
\small
\caption{Where a metadata-only DMR-IR score comes from, on the 236 patients and the same
patient-level folds as the imaging arms. \emph{Tabular} rows are logistic models over field
subsets; \emph{Text} rows are TF--IDF probes over clause ablations of the distributed
prompt. \emph{Field presence only} replaces each value by whether it was recorded.}
\label{tab:metadata}
\begin{tabular}{lcc}
\toprule
Subset & Repr. & AUROC \\
\midrule
Pre-diagnostic fields & Tabular & 0.668 [0.595, 0.737] \\
Care-pathway fields & Tabular & 0.811 [0.753, 0.866] \\
\texttt{biopsy} alone & Tabular & 0.714 [0.639, 0.785] \\
Field presence only & Tabular & 0.555 [0.478, 0.631] \\
All fields & Tabular & 0.952 [0.921, 0.976] \\
\midrule
Full prompt & Text & 0.927 [0.888, 0.960] \\
$-$ care-pathway clause & Text & 0.908 [0.864, 0.944] \\
Care-pathway clause only & Text & 0.855 [0.802, 0.900] \\
Pre-diagnostic clauses & Text & 0.899 [0.854, 0.938] \\
\bottomrule
\end{tabular}
\end{table}

\FloatBarrier
\section{Cross-cohort transfer (not load-bearing)}
\label{sec:transfer}
Zero-shot DMR-IR $\to$ second cohort: 0.798 [0.728, 0.860] in-domain falls to
0.519 [0.410, 0.624] (per-image standardisation) and 0.775 [0.696, 0.846] to
0.457 [0.351, 0.564] (fixed temperature scale). The in-domain references are the
frozen frontal ConvNeXt-T probe scored on its own held-out source patients
\emph{within the transfer pipeline}, on one source fold and under the
normalisation of the transfer arm each is paired with, which is why they read
0.798 and 0.775
rather than the 0.815/0.811 of the five-repeat cross-validated arm; each is the
correct reference for the drop beside it. Both transfer intervals contain 0.5. The two
cohorts differ in camera, country, population, protocol, geometry and label
semantics; a frozen probe is least able to absorb that shift, so this result
does not by itself distinguish absent signal from unadapted signal and is
reported only as convergent context.

\FloatBarrier
\section{Model selection on a small validation partition}
\label{sec:selection}
An initial fine-tuned run selected the stopping epoch by maximising AUROC on a
15-patient (4 positive) validation partition, whose AUROC moves in increments of
$1/44$. Selected-validation vs.\ test AUROC correlated at $+0.06$ across 25
folds, mean optimism 0.14, median selected epoch 4, and 14/25 folds emitted a
near-constant probability. The run was discarded on this validity criterion
(never referencing test labels) and retained with diagnostics; the four are
recomputed from its stored fold log and predictions by
\path{scripts/43_selection_diagnostics.py}.

The scope of epoch selection is as follows. Validation-based epoch selection was
removed from every \emph{fine-tuned} arm and from every rung of the matched
ladder: those runs train to a fixed budget and are scored at its end
(\mbox{\texttt{early\_stopping: false}} in their stored \texttt{config.json}). The
frozen linear probes keep it, restoring the weights of the epoch with the
best validation AUROC and using patience, and that includes the frontal probe and
both all-frames arms of Table~\ref{tab:channels}. This is the one difference
between the 0.815 of that table and the 0.833 of the ladder's reference rung: the
same encoder, the same view, the same patients, selected epoch against fixed
budget. Nothing in the paper compares a selecting arm with a fixed-budget arm as
though they were the same configuration.

\FloatBarrier
\section{Age on DMR-IR: a null, not an inversion}
\label{sec:age}
Table~\ref{tab:channels} scores age through a logistic model fitted on the same
folds as the image, so that the two are comparable, and that arm reads 0.416,
below chance, with an interval that excludes it. Read on its own, a sub-chance
AUROC points at an inverse association, and reversing the score would give
0.584. The covariate scored directly says otherwise: age reaches 0.495 [0.415,
0.576] against the diagnosis on the same 236 patients, on mean ages of 54.8
(benign) and 54.5 (malignant) years (\path{scripts/42_age_direction.py}).
There is no association to invert.

A fitted arm can sit reliably below chance where a raw one sits at chance
because cross-validation folds partition one finite cohort: with no signal to
fit, the coefficient estimated on the training fold reflects that fold's noise,
which is complementary to the held-out fold's. The fit then ranks slightly
against the test fold, run after run. That is why the paper reports the raw
statistic when the question is direction, and the fitted one when the question
is what a model of that channel would achieve beside a model of the image. The
same script scores the second cohort at 0.769 [0.676, 0.850] on mean ages of
45.1 and 57.4 years, reproducing the raw value quoted in \S\ref{sec:second}.

\FloatBarrier
\section{Exploitation on the second cohort}
\label{sec:exploitation2}
The same statistic as Table~\ref{tab:exploit_all}, computed by the same estimators:
\path{scripts/37_verification_confounder.py} imports the bootstrap and permutation
routines from \path{scripts/34_exploitation_interval.py} rather than reimplementing
them, so the two cohorts' columns are numerically comparable. The confounder here is
continuous, so each diagnosis stratum is split at its own median age, within the stratum and
never across the cohort, since malignant patients are twelve years older and a
cohort-wide split would reproduce the label. Patients sitting exactly on the
median are excluded rather than assigned, which leaves 79 of the cohort's 84
benign patients (5 sit on the benign median of 45 years) and 34 of its 35
malignant ones (1 sits on the malignant median of 59). The two strata are
analysed separately and are never pooled; the cohort itself is 119 patients
throughout the paper.

Every arm with diagnostic discrimination above chance carries age into its score within a
diagnosis; the frozen probe, which has none, does not (Table~\ref{tab:exploitation2}).
Unlike DMR-IR's calendar, age here is a legitimate clinical predictor: the licensed reading
is that these arms track an age-correlated appearance, which is why they add nothing over
age, not that the benchmark is invalid.

\begin{table}[h]
\centering\small
\caption{Within-diagnosis age discrimination from each arm's own malignancy score, on the
second cohort. Strata are split at their own median age, leaving 79
benign and 34 malignant patients. \emph{Spearman} scores the same benign patients against
untransformed age and does not depend on where the split falls. The within-malignant column
is reported for completeness on 34 patients and is not used. The permutation $p$-values
after each interval are uncorrected; across the three configurations that pass the quality
gate, the benign Spearman correlations of the two fine-tuned arms survive a Holm correction
and those $p$-values do not. $\dagger$: configuration
failing the pre-specified quality gate. $\ddagger$: a second run of the row below it.}
\label{tab:exploitation2}
\begin{tabular}{lcccc}
\toprule
Arm & Diagn. & Age within benign & Spearman & Age within \\
 & AUROC & AUROC [95\% CI], $p$ & (benign) & malignant \\
\midrule
Swin-T, fine-tuned & 0.669 & 0.630 [0.504, 0.760], 0.048 & $+0.35$ & 0.830 \\
ConvNeXt-T, fine-tuned$^{\ddagger}$ & 0.658 & 0.642 [0.513, 0.765], 0.030 & $+0.29$ & 0.806 \\
ConvNeXt-T, fine-tuned & 0.654 & 0.641 [0.510, 0.764], 0.030 & $+0.29$ & 0.824 \\
ResNet-50, fine-tuned$^{\dagger}$ & 0.581 & 0.589 [0.455, 0.720], 0.18 & $+0.27$ & 0.668 \\
ConvNeXt-T, frozen & 0.535 & 0.501 [0.370, 0.630], 0.99 & $-0.03$ & 0.633 \\
\bottomrule
\end{tabular}
\end{table}

The within-malignant column carries no claim: with 17 patients per group its intervals are
too wide to separate an effect from its absence, and the stratum is where the cohort's own
age spread is smallest.

The legibility question was then asked directly, as on DMR-IR: a probe trained
to recover the within-stratum age band from the pixels, on the same folds
(\path{results/experiment_ver_12_age_*}). It does not succeed. Within the
79 benign patients the ConvNeXt-T probe reaches 0.476 [0.344, 0.608], and within
the 34 malignant patients 0.602 [0.394, 0.804]; both intervals contain chance.
Both Swin-T arms fail the same prediction-spread gate that ResNet-50 fails on
DMR-IR: the within-benign arm collapsed on 9 of 25
folds and the within-malignant arm on 8, against the 5 the gate allows, and
both are reported for transparency rather than as evidence either way.

This is the mirror image of DMR-IR and it is reported as such: there the
confounder was legible to every encoder and demonstrably used by one arm of the
seven that pass the quality gate; here it is demonstrably present in the diagnostic scores and \emph{not}
demonstrably legible to a probe. We do not claim age is absent from the pixels
since 79 benign and 34 malignant patients split at their own medians is a weak
test, and a null on it is not evidence of absence. What the pair of results does show is
that the two
measurements are not substitutes for one another in either direction, which is
the methodological point of running both.

\FloatBarrier
\section{Matching diagnostics}
\label{sec:balance}
Standardised mean differences for every rung of both ladders, computed from the
\texttt{balance.csv} each cohort build writes and tabulated by
\path{scripts/38_matching_balance_report.py}. Two readings matter. The confounder's
own row must
approach zero on the rungs that control it: the day calipers reach $|$SMD$|
\leq 0.05$ while the overlap rung only falls from $+2.46$ to $+1.20$, which is
the arithmetic behind describing that rung as weakening rather than removing the
confounder. And the null control must leave it alone: propensity matching on
clinical covariates moves acquisition from $+2.46$ to $+2.39$. Size-control
draws are omitted, because their balance describes a random draw rather than a
design, and are in \path{results/matching/smd_all_rungs.csv}.

\begin{table}[h]
\centering
\small
\caption{DMR-IR: standardised mean difference (malignant minus benign over the pooled
standard deviation) per covariate, one row per rung of Table~\ref{tab:ladder}.}
\label{tab:smd_dmr}
\begin{tabular}{lcccc}
\toprule
Cohort & acquisition\_days & age\_at\_visit & body\_temp. & menarche \\
\midrule
\texttt{original} & $+2.46$ & $-0.02$ & $+0.16$ & $-0.25$ \\
\texttt{overlap} & $+1.20$ & $+0.15$ & $+0.04$ & $-0.06$ \\
\texttt{year\_balanced} & $+0.39$ & $+0.16$ & $+0.83$ & $-0.22$ \\
\texttt{days\_matched\_7} & $+0.01$ & $-0.25$ & $+0.93$ & $+0.67$ \\
\texttt{days\_matched\_30} & $-0.01$ & $-0.09$ & $+0.63$ & $+0.74$ \\
\texttt{days\_matched\_90} & $+0.05$ & $+0.08$ & $+0.46$ & $+0.73$ \\
\texttt{days\_matched\_180} & $-0.04$ & $+0.39$ & $-0.33$ & $+0.05$ \\
\texttt{propensity\_clinical} & $+2.39$ & $-0.17$ & $+0.09$ & $+0.09$ \\
\texttt{propensity\_w\_acq} & $-0.09$ & $+0.27$ & $+0.67$ & $+0.08$ \\
\bottomrule
\end{tabular}
\end{table}

Controlling the acquisition axis costs balance elsewhere: the narrow calipers retain 20--26
patients and their body-temperature and menarche SMDs rise accordingly, which is one more
reason those rungs are reported as descriptive only. The second cohort's age rungs balance
the confounder without disturbing BMI or body temperature (Table~\ref{tab:smd_ver}): a
cleaner ladder than DMR-IR's, which is what a cohort with a single centre, camera and
three-year window affords.

\begin{table}[h]
\centering
\small
\caption{Second cohort: standardised mean difference per covariate, same convention as
Table~\ref{tab:smd_dmr}, one row per rung of its age ladder.}
\label{tab:smd_ver}
\begin{tabular}{lccc}
\toprule
Cohort & age & bmi & temperature \\
\midrule
\texttt{original} & $+1.05$ & $+0.33$ & $+0.14$ \\
\texttt{overlap} & $+0.56$ & $+0.24$ & $+0.19$ \\
\texttt{age\_matched\_2} & $-0.02$ & $-0.07$ & $+0.01$ \\
\texttt{age\_matched\_5} & $+0.04$ & $-0.05$ & $+0.05$ \\
\texttt{age\_matched\_10} & $+0.20$ & $-0.04$ & $+0.03$ \\
\texttt{propensity\_clinical} & $+1.00$ & $+0.02$ & $-0.03$ \\
\texttt{propensity\_with\_age} & $-0.01$ & $+0.11$ & $+0.03$ \\
\bottomrule
\end{tabular}
\end{table}

\FloatBarrier
\section{Second-cohort description}
\label{sec:cohort2}
The cohort holds 119 patients, 84 benign and 35 malignant. No subject is
bilaterally normal, so the contrast is malignancy against benign breast disease in
referred women, and malignancy is exclusively unilateral. Contralateral-asymmetry
descriptors reach only 0.638 for a structural reason: every patient has unilateral
pathology, so asymmetry measures a property both classes share. Among the
covariates, age separates the classes ($r_{pb}=+0.428$, $p=1.2\times10^{-6}$);
weight, height, BMI and body temperature do not (all $p>0.1$).

\FloatBarrier
\section{Source of the acquisition-era signature in the thermograms}
\label{sec:signature}
The era is legible to every encoder (0.813--0.823), which says nothing about
\emph{what} is legible. Because the mirror distributes a breast mask with every
frame, the question can be answered without a network: the same within-benign
era contrast, on the same patient-level folds and bootstrap, decoded by a
standardised logistic regression from four tabular descriptions of one static
frontal frame per patient (\path{scripts/39_signature_source.py}).

\begin{table}[h]
\centering\small
\caption{Acquisition era within the 133 benign patients, by what the decoder is allowed to
see. The percentile blocks are mean, standard deviation, min, max, five percentiles, IQR,
skewness and kurtosis of the selected pixels; the histogram is 32 bins fixed at
18--38$^{\circ}$C plus the fraction of pixels inside that range, so a bin means the same
temperature in every image. Masks are upsampled to the frame rather than the frame
downsampled to the mask.}
\label{tab:signature}
\begin{tabular}{lcc}
\toprule
Representation & Features & Era AUROC [95\% CI] \\
\midrule
Background only & 12 & 0.938 [0.892, 0.976] \\
Whole frame & 12 & 0.924 [0.875, 0.966] \\
Temperature histogram & 33 & 0.899 [0.841, 0.951] \\
Breast region only & 12 & 0.806 [0.729, 0.875] \\
\bottomrule
\end{tabular}
\end{table}

Three readings follow, and none of them needs a model. The signature is
\emph{outside} the breast: background pixels alone beat every encoder in
Table~\ref{tab:encoders}, while the breast region alone is the weakest of the four. It
needs no spatial structure: a histogram of the frame, with all geometry discarded,
reaches 0.899. And it is a shift in absolute temperature, since these are
degrees Celsius recovered from the radiometric files rather than display values.
What separates 2012 from 2013 on this benchmark is, to a first approximation,
how warm the whole frame was, background included: the trace a change in the
acquisition environment would leave, although DMR-IR does not record that
environment and so cannot establish the cause (\S\ref{sec:legibility}). That is
the most literal form of the acquisition confounding this paper is about, and the
least plausible candidate for a pathological signal. The breast region's own 0.806 does not contradict
this: a global thermal offset moves the breast with everything else.

\FloatBarrier
\section{Era legibility under fine-tuning}
\label{sec:finetuneera}
The era column of Table~\ref{tab:encoders} is a frozen-representation decode, while the
strongest frontal diagnostic arms are fine-tuned. To ask whether fine-tuning for
malignancy changes what the representation holds, the two fine-tuned frontal
arms were rerun with their penultimate-layer activations retained
(\path{scripts/04_train_single_view.py --save-features}), together with the
frozen arms of the same encoders, and an identical logistic probe of the era was
fitted \emph{inside each fold}, on that fold's training patients, scored on
its held-out patients and pooled out of fold
(\path{scripts/40_finetuned_era_probe.py}). Twenty-five fine-tuned folds are
twenty-five different feature spaces, so a single probe over pooled activations
would average over incomparable bases. The reruns reproduce the arms they
duplicate, fine-tuned (ConvNeXt-T 0.9230 against 0.9228; Swin-T 0.9187 against 0.9199)
and frozen (ConvNeXt-T 0.8149 against 0.8149; Swin-T 0.8737 against 0.8723), which is
the check that the activation dump did not disturb training.

\begin{table}[h]
\centering\small
\caption{Acquisition-era AUROC within the 133 benign patients, decoded from the
representation of the diagnostic model, frozen versus fine-tuned. Same probe, same folds,
same patients, so the difference is paired.}
\label{tab:finetuneera}
\begin{tabular}{lccc}
\toprule
Encoder & Frozen & Fine-tuned & Paired $\Delta$ [95\% CI] \\
\midrule
ConvNeXt-T & 0.953 & 0.945 & $-0.008$ [$-0.027$, $+0.007$] \\
Swin-T & 0.937 & 0.940 & $+0.003$ [$-0.023$, $+0.031$] \\
\bottomrule
\end{tabular}
\end{table}

Two things follow. Training for malignancy neither sharpens nor erases the
acquisition signature: both differences are small and their intervals contain
zero. And the absolute level, 0.94--0.95, sits well above the 0.813--0.823 of
Table~\ref{tab:encoders}, not because the representation differs, since the frozen
row here \emph{is} that representation, but because a logistic probe on pooled
penultimate features reads it better than a linear head trained end-to-end on
the same features does. The paper's legibility numbers are therefore a
conservative reading of how much era information a representation carries. That
does not weaken any claim in the body of the paper; it means the claims are made against
the lower of two available estimates. It also matches Appendix~\ref{sec:signature}: a
signature that is essentially a global temperature offset is exactly the kind a
simple linear read-out recovers.

\end{document}